\documentclass[american]{article}
\PassOptionsToPackage{natbib=true}{biblatex}
\usepackage[T1]{fontenc}
\usepackage[utf8]{inputenc}
\usepackage{babel}
\usepackage{array}
\usepackage{varwidth}
\usepackage{graphicx}
\usepackage{geometry}
\usepackage[pdfusetitle,
 bookmarks=true,bookmarksnumbered=false,bookmarksopen=false,
 breaklinks=true,pdfborder={0 0 1},backref=false,colorlinks=false]
 {hyperref}

\makeatletter

\providecommand{\tabularnewline}{\\}

\makeatother

\usepackage[style=authoryear,backend=bibtex]{biblatex}
\begin{document}
\title{Toward a systematic method for identifying language areas}
\author{Hiram Ring\\
NTU Singapore}
\maketitle
\begin{abstract}
Macroareas are geographical areas used in typological research for
grouping variables of interest. In linguistic typology, languages
in a given macroarea are considered to have potential for contact,
in contrast to those outside the area, where contact is less likely.
Along with language family membership, macroareas are used as controls
for models in linguistic typology, in an attempt to address the problem
of autocorrelation - the observation that historical developments
or typological patterns may be due to contact between neighboring
languages and/or inheritance from a common ancestral language. Macroareas
are therefore a central aspect of research that seeks to separate
universal properties of language from local (or language-specific)
properties. Existing macroareas largely depend on expert determinations
of what constitutes a geographical area of potential contact, and
to date have mainly aligned with continents or landmasses \citep{Hammarstrom:2014aa,Nichols:2013aa}.
While there are various historical and theoretical reasons for these
groupings, there as of yet has been no systematic approach to identifying
such areas for a given region. This paper attempts to address such
a gap and move beyond macroarea to identification of language areas
of relatively arbitrary size, presenting a simple geographical clustering
method for identifying groupings over any area. The method produces
a set of worldwide macroareas that largely align with existing groupings,
as well as local groupings for a well-known sprachbund.
\end{abstract}

\section{Introduction}

The discipline of linguistic typology is concerned with observing
patterns among languages of the world for the purposes of comparison.
Linguists engaged in typological research therefore often interact
with multiple other disciplines, making use of grammatical descriptions
for particular languages to develop comparative datasets, using theoretical
constructs to organize such datasets, adapting tools and methods from
related fields for analysis, and drawing inferences from comparison
of the relevant data based on insights from human behavior. Typological
comparisons have led to the proposal of universal properties of language
for various domains, such as phonology \citep{Hyman:2006aa,Parker:2026aa},
morphology \citep{Spencer:2006aa}, and syntax (word order \citep{Dryer:1992fk,Bickel:2011aa,Croft:2011aa}),
as well as the identification of possible universal pressures that
shape language change \citep{Fedzechkina:2011aa,Gell-Mann:2011aa,Hahn:2020aa,Hawkins:2014aa,Ring:2025ab}.

From its inception, linguistic typology has largely focused on data
gathering and dataset development, which is understandable given the
dearth of data for the majority of the world's 7,000+ languages. The
increasing availability of larger datasets has supported a trend toward
analysis, with recent discussion being focused on methodological concerns
and best practices for evaluating properties of language in such datasets
\citep{Guzman-Naranjo:2022aa,Becker:2025aa,Miestamo:2025aa,Serzant:2025aa,Becker:2025ab,Shcherbakova:2025aa}.
Along with a greater emphasis on computational and statistical methodologies,
a set of best practices are gradually emerging regarding dataset development,
language sampling, and control variables, supporting researchers to
adequately model and measure typological features of interest.

\begin{table}
\begin{centering}
\begin{tabular}{|llV{\linewidth}|}
\hline 
Domain &  & Best practices\tabularnewline
\hline 
 &  & \tabularnewline
Datasets: &  & \begin{enumerate}
\item link or reference underlying data
\item make data accessible
\end{enumerate}
\tabularnewline
 &  & \tabularnewline
Language samples: &  & \begin{enumerate}
\item as large as possible
\item distribution reflects context
\end{enumerate}
\tabularnewline
 &  & \tabularnewline
Controls: &  & \begin{enumerate}
\item language area, family
\item applied consistently
\end{enumerate}
\tabularnewline
 &  & \tabularnewline
\hline 
\end{tabular}
\par\end{centering}
\caption{Best practices for typological linguistic research}\label{tab:Best-practices-for}

\end{table}

As an example of best practices, datasets used for typological research
can be extensive or narrow, but should link to or reference underlying
data, which ideally would be accessible in some way. Language samples
should be as large as possible for a given context, with a distribution
that reflects the context of investigation as well as possible. Controls
for known effects - minimally language area and language family -
should be used when conducting statistical tests on correlations.
These best practices, by no means an exhaustive list, are summarized
in table \ref{tab:Best-practices-for}. While it may not be possible
to meet all these criteria in every study, such practices are crucial
for linguistic typology to make claims regarding universal properties
of language.

It has long been recognized that typological claims may be undermined
by the impact of autocorrelation. This is also known as ``Galton's
problem'' \citep{Bromham:2024aa}, i.e. the observation that typological
and other similarities between languages may be a result of languages
sharing a common ancestor (descent) or of contact effects between
languages spoken in more or less contiguous geographical areas (language
area). Due to the history of human migration and habitation, these
two concerns are often strongly collinear, though they are typically
controlled for separately in statistical models. To deal with the
issue of descent, a categorical variable for family membership or
family tree is typically used, though recent research also uses a
hierarchical phylogenetic tree \citep{Verkerk:2025aa}. To deal with
the issue of language area, categorical variables that group neighboring
languages are often used, possibly alongside a gaussian process that
incorporates geographical location (latitude, longitude). The remainder
of this paper focuses on a particular control (``macroarea'') proposed
to deal with one aspect of the problem of autocorrelation in language
comparison.

\section{Macroareas and clustering algorithms}

The concept of ``macroareas'' was developed to provide a control
for typological observations involving a worldwide sample of languages.
Existing macroareas largely depend on expert determinations of what
constitutes a geographical area of potential contact, and to date
have mainly aligned with continents or landmasses \citep{Hammarstrom:2014aa,Nichols:2013aa}.
While there are various historical and theoretical reasons for these
groupings, the existing systematic approach to identifying such areas
for a given region involves fully manual annotation.

Besides requiring extensive knowledge regarding the languages and
areas in question, manual approaches are time-consuming and can suffer
from human error. A case in point regards the annotation of the language
``Chinese Pidgin English'' as belonging to the ``Eurasian'' macroarea
in Glottolog's database, despite being located on the island of Nauru
in the Pacific Ocean (and note comments in \citealt{Hammarstrom:2014aa}
regarding problems with the WALS macroareas). Additionally, such areas
are somewhat arbitrary in the sense that researchers must justify
decisions regarding the number and size of macroareas, as well as
the boundaries. Although this is true to some degree with any approach,
I propose that rather than depend on fully manual annotation of membership
in one or another area, researchers can make use of mathematical clustering
algorithms for geographical coordinates to assist in developing principled
areas for comparison. As a final point, previous approaches cannot
be easily applied to areas of varying sizes, as opposed to the approach
I propose.

\subsection{Geographical coordinate representation}

Geographical coordinates locate points on the globe based on latitude
and longitude, which can also be used to mark the locations where
languages are spoken. As continuous variables, latitude and longitude
also allow researchers to identify distance between points. This is
the primary reason for including coordinates as input for a gaussian
process term in typological studies \citep{Guzman-Naranjo:2022aa,Guzman-Naranjo:2024aa,Verkerk:2025aa}.
Distance metrics commonly used in geospatial study include Hilbert
distance \citep{Lei:2023aa} and Euclidean or Cartesian distance \citep{Guzman-Naranjo:2024aa}.
While both metrics preserve locality well, Hilbert values are one-dimensional
transformations of coordinates primarily used to improve efficiency
in database queries, while Euclidean distance is more widely used
in most applications for its low complexity and ease of implementation
while preserving two-dimensional information. Additionally, Hilbert
values can lead to geographical ``discontinuity errors'' \citep{Rauscher:2025aa}
where neighboring points are placed far apart or grouped separately
due to their placement in the one-dimensional sequence, which makes
Euclidean distance a better choice for most applications.

\subsection{Deriving macroareas using clustering}

Clustering is a statistical approach to identifying patterns in data.
It is unsupervised, which means that it is agnostic to the properties
of the dataset. Various methods for clustering seek to identify similarities
between datapoints in order to find the best ``cluster'' or ``split''
of the data into different groups. Methods that are widely used in
geospatial science applications include K-Means, K-Mediods, Silhouette
and Heirarchical Clustering \citep{Sadeghi:2025aa}. K-Means is one
of the more computationally efficient and simple algorithms to use,
which makes it a good choice for our goal of identifying clusters
among languages. For evaluation, there are similarly multiple techniques,
including Silhouette, Davies-Bouldin, Calinski-Harabasz, Elbow, and
Bayesian Information Criterion. For ease of interpretation, the Silhouette
score is often preferred.

\subsection{Worldwide: K-means clusters vs Glottolog macroareas}

\begin{figure}
\includegraphics[width=0.9\columnwidth]{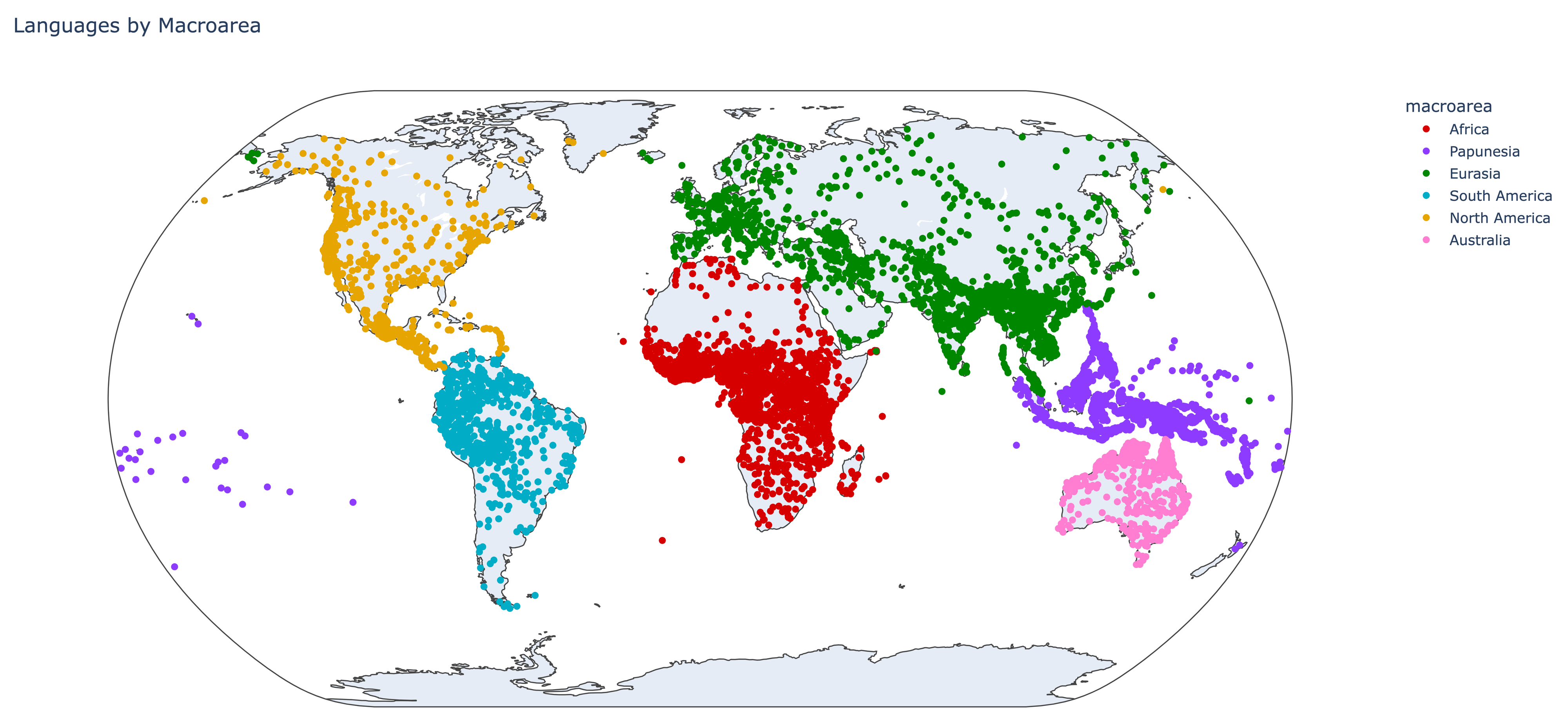}

\caption{Macroareas from Glottolog}\label{fig:Macroareas-from-Glottolog}

\end{figure}

To identify macroareas via clustering, I downloaded the Glottolog
V5.3 data on languages and dialects.\footnote{https://glottolog.org/meta/downloads ``languages\_and\_dialects\_geo.csv''}
This file contains over 22,000 glottocodes, of which 8,889 `languoids'
contain latitude and longitude coordinates. These points were then
plotted on a map (figure \ref{fig:Macroareas-from-Glottolog}).\footnote{For Python code related to this see: https://www.hiramring.com/blog/2026-07-28-Language\_areas\_clustering.html}

\begin{figure}
\includegraphics[width=0.9\columnwidth]{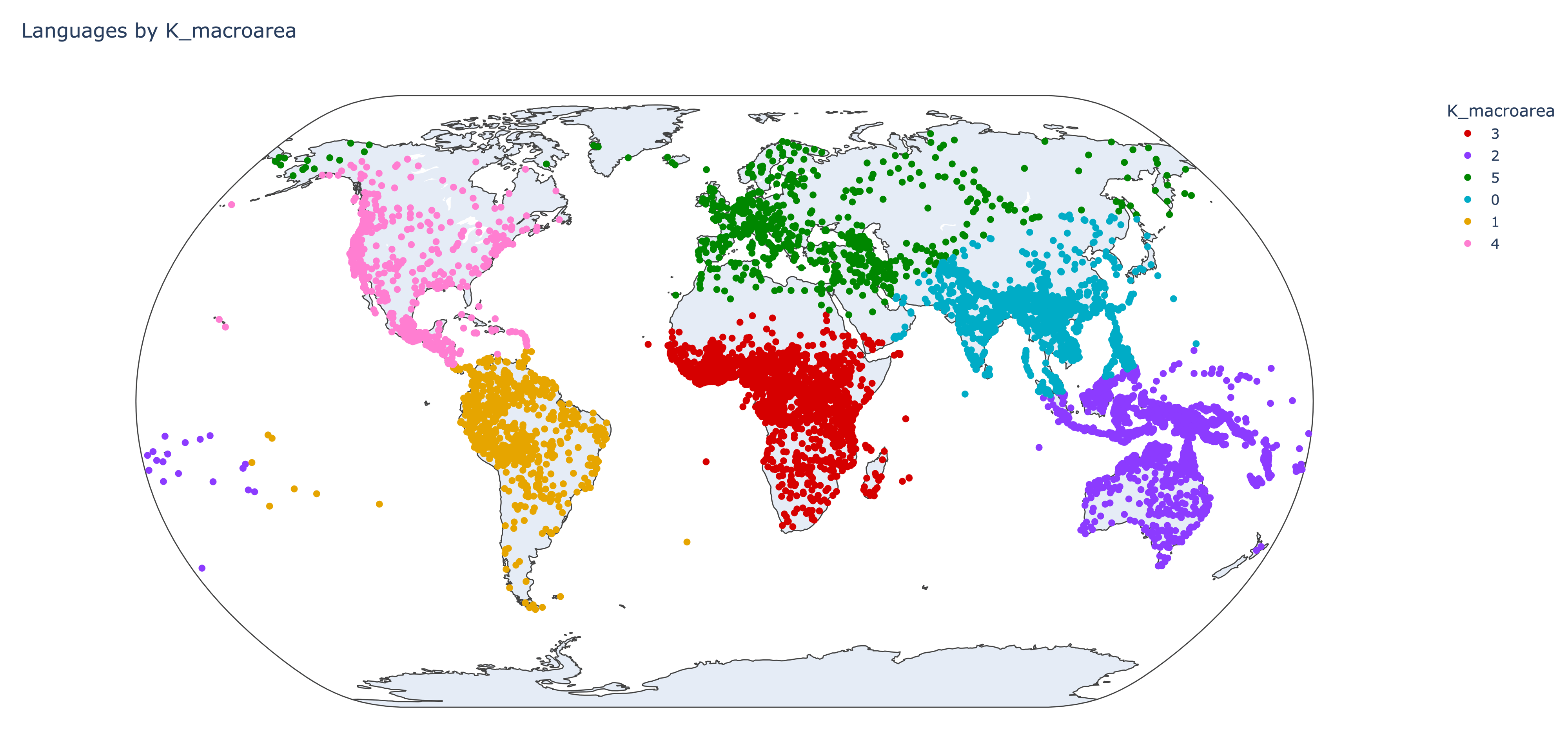}

\caption{Macroareas from K-means}\label{fig:Macroareas-from-K-means}
\end{figure}

After converting longitude and latitude to three-dimensional Cartesian
space, I used the \emph{scikit-learn} Python library to cluster the
language points for \emph{k = range(2, 20)} and return a silhouette
score for each \emph{k}. The \emph{k} with the highest score was considered
the optimal number of clusters. This resulted in 6 macroareas identified,
which were also plotted on a map (figure \ref{fig:Macroareas-from-K-means}).

\subsection{Smaller sprachbunds: K-means clusters vs local areas}

\begin{figure}
\includegraphics[width=0.9\columnwidth]{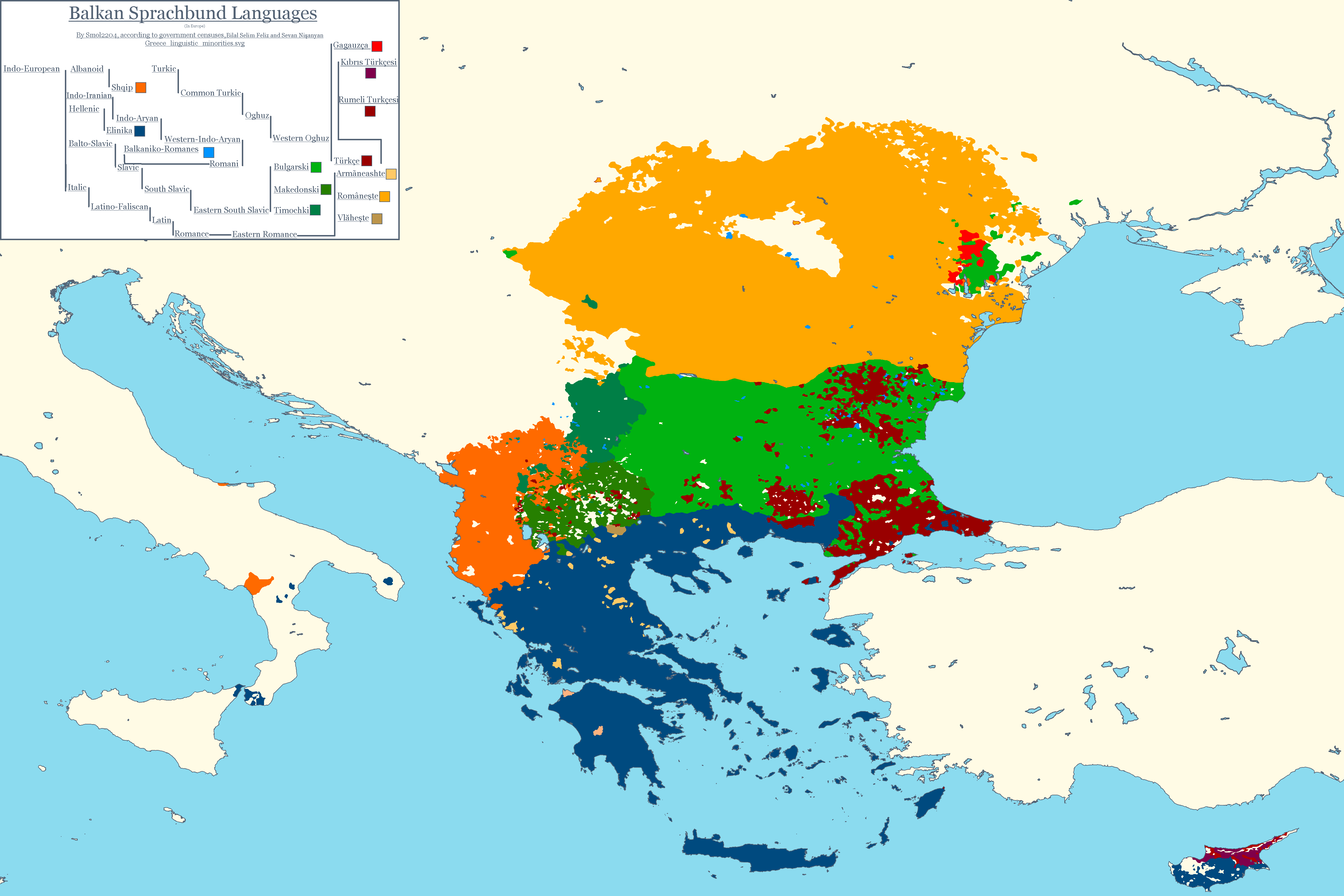}

\caption{Balkans sprachbund languages (Wikimedia commons)}\label{fig:Balkans-sprachbund-languages}
\end{figure}

In theory, the clustering methodology should work with any bounded
area. To test this, I took latitude and longitude coordinates that
surround the Balkan sprachbund (a well-known language contact area)\footnote{Southwest: (32.90056179070983, 7.405988927677103), Northeast: (49.71221586095594,
32.63059823530507)} and reduced the Glottolog dataset to only those languages found between
those coordinates. A map of languages of the Balkans is presented
as figure \ref{fig:Balkans-sprachbund-languages}.\footnote{https://upload.wikimedia.org/wikipedia/commons/d/df/Languages\_of\_the\_Balkan\_Sprachbund\_in\_the\_Balkans\%2C\_Cyprus\_and\_Italy.png}

\begin{figure}
\includegraphics[width=0.9\columnwidth]{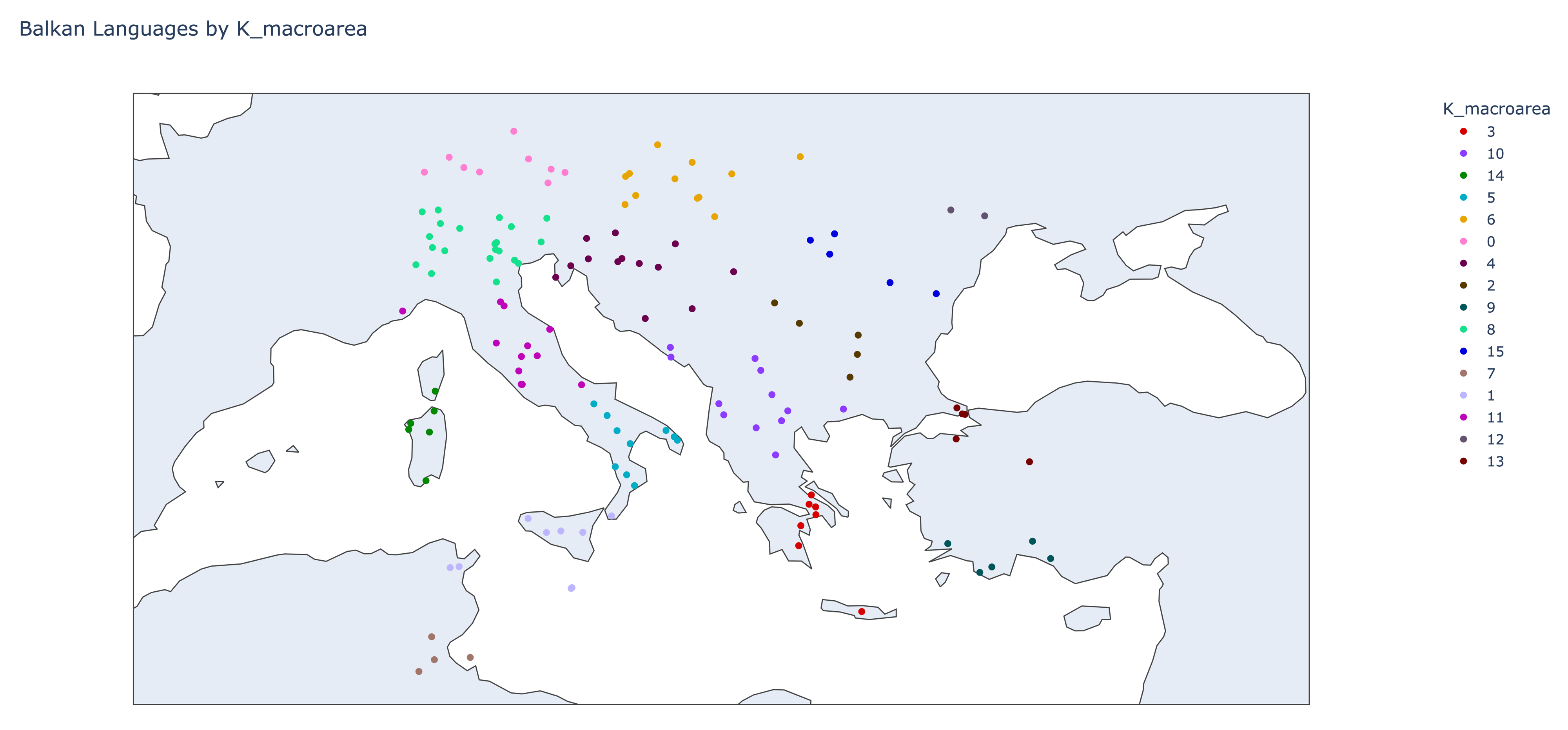}

\caption{K-means clusters of Balkans sprachbund}\label{fig:K-means-clusters-of}
\end{figure}

I then used the same conversion and clustering methodology as for
the global dataset, which resulted (for this area) in a \emph{k} of
16. The languages were plotted as in figure \ref{fig:K-means-clusters-of}.

\section{Discussion}

There are several points worth noting regarding this exercise, the
first of which is that the worldwide K-means clusters derived from
the geographical coordinates are strikingly similar to the Glottolog
macroareas. The differences between the two groupings actually raise
questions regarding the correct way to group languages in relation
to geography - the K-means clusters seem to divide areas better at
certain geographical areas such as the Himalayan mountains and the
Wallace line, which have been posited as barriers to human migration.
The similarities indicate that expert determinations of \textquotedbl language
area\textquotedbl{} align quite well with a more naive mathematical
approach. One major open question, however, is how well such K-means
clusters serve as controls for other typological investigations.

The clustering of languages in the Balkan area is somewhat less striking,
in the sense that it is hard to know whether the identified clusters
indicate ``language areas'' in the sense traditionally understood
by linguists. Since this is only a clustering based on geographical
location, it is agnostic to input, but at the same time sensitive
to such input. For example, we do not know for certain whether the
points identified for a given language represent the center of where
the language is spoken, or an outlying area. It may be that using
polygons instead of points, or simply increasing the observed area
(increasing the ``resolution'' in some way) would adjust this outcome.

It seems clear, however, that there are some benefits to using an
unsupervised clustering method for identifying potential geographic
relationships between languages. Given a large number of datapoints,
we are able to observe trends and patterns that can inform our analysis.
This allows us to make better decisions about what languages \emph{should}
belong to a given macroarea, and refer to external sources for verification.
I would suggest, for example, on the basis of this excercise, that
the Glottolog macroareas could be adjusted to account for the Himalayas
and the Wallace Line, and that Australia could be separated in the
K-means clusters.

Whether such adjustments would lead to different analyses or findings
is an empirical question for future work. The general observation,
however, is that the tools we have for exploring data via clustering
and mapping should be used. In the case of smaller language areas,
the method described here can be a good starting point for understanding
the patterns visible due to geographical proximity, which have been
central to linguistic contact and human interaction.

\printbibliography

\end{document}